\documentclass[10pt,twocolumn,letterpaper]{article}

\usepackage{iccv}
\iccvfinalcopy
\usepackage{times}
\usepackage{epsfig}
\usepackage{graphicx}
\usepackage{amsmath}
\usepackage{amssymb}

\usepackage{booktabs}
\usepackage{amsfonts}
\usepackage{algorithm}
\usepackage{algorithmicx}
\usepackage{algpseudocode}
\usepackage{subfigure}
\usepackage{multirow}
\usepackage{diagbox}
\usepackage{amsthm}
\usepackage{bm}
\usepackage[T1]{fontenc}
\renewcommand{\mathbf}{\boldsymbol}

\def\x{\mathbf{x}}

\def\u{\mathbf{u}}

\def\I{\mathbf{I}}

\DeclareMathOperator*{\argmax}{arg\,max}
\DeclareMathOperator*{\argmin}{arg\,min}

\usepackage{dsfont}

\usepackage[pagebackref=true,breaklinks=true,letterpaper=true,colorlinks,bookmarks=false]{hyperref}

\def\iccvPaperID{5653} 

\ificcvfinal\fi

\begin{document}

\title{Unsupervised Domain Adaptation of Black-Box Source Models}

\author{Haojian Zhang$^1$, Yabin Zhang$^2$, Kui Jia$^1$, Lei Zhang$^2$\\
	$^{1}$South China University of Technology \quad
	$^{2}$The Hong Kong Polytechnic University \\
	{\tt\small eehjzhang@mail.scut.edu.cn,csybzhang@comp.polyu.edu.hk, }\\ {\tt\small kuijia@scut.edu.cn,cslzhang@comp.polyu.edu.hk}
}

\maketitle
\ificcvfinal\thispagestyle{empty}\fi

\begin{abstract}
	Unsupervised domain adaptation (UDA) aims to learn models for a target domain of unlabeled data by transferring knowledge from a labeled source domain. In the traditional UDA setting, labeled source data are assumed to be available for adaptation. Due to increasing concerns for data privacy, source-free UDA is highly appreciated as a new UDA setting, where only a trained source model is assumed to be available, while labeled source data remain private. However, trained source models may also be unavailable in practice since source models may have commercial values and exposing source models brings risks to the source domain, e.g., problems of model misuse and white-box attacks. In this work, we study a subtly different setting, named Black-Box Unsupervised Domain Adaptation (B$^2$UDA), where only the application programming interface of source model is accessible to the target domain; in other words, the source model itself is kept as a black-box one. To tackle B$^2$UDA, we propose a simple yet effective method, termed Iterative Learning with Noisy Labels (IterLNL).  With black-box models as tools of noisy labeling, IterLNL conducts noisy labeling and learning with noisy labels (LNL), iteratively. To facilitate the implementation of LNL in B$^2$UDA, we estimate the noise rate from model predictions of unlabeled target data and propose category-wise sampling to tackle the unbalanced label noise among categories. Experiments on benchmark datasets show the efficacy of IterLNL. Given neither source data nor source models, IterLNL performs comparably with traditional UDA methods that make full use of labeled source data.
\end{abstract}

\section{Introduction}
\label{SecIntro}

Although deep models have achieved success on various tasks, it is difficult to generalize the model learned from labeled training data to a target domain of slightly shifted data distribution. At the same time, it is expensive to collect a new target dataset with a large number of labeled training data. Therefore, unsupervised domain adaptation (UDA) \cite{transfer_survey,dan,dann} is introduced to learn the target model by transferring knowledge from the labeled source domain to the unlabeled target domain.
Motivated by seminal theories \cite{ben2010theory,zhang2019bridging},  popular UDA methods \cite{dan,dann,mcd, cada, symnets} target at learning domain invariant feature representations.
The underlying motivation is that the source classifier could be safely applied to the target domain once domain invariant feature representations are achieved. In UDA, labeled source data are assumed to be available for target domain.

Although remarkable success has been achieved in UDA, increasing concerns for data privacy post new challenges to this task. Specifically, data of source and target domains are typically captured and stored on different devices and contain private information. Thus it is risky to expose source data to the target domain and vice versa. In other words, labeled source data may be not available for the target domain, impeding the application of popular UDA methods \cite{dan, dann, mcd, cada, symnets}. For this reason, a novel task, source-free UDA, is introduced \cite{yang2020unsupervised, hou2020source, liang2020we} to facilitate the model adaptation and protect the source data privacy simultaneously.

Unlike the vanilla UDA, a well-trained source model, instead of labeled source data, is provided to unlabeled target domain in the source-free UDA \cite{yang2020unsupervised, liang2020we}.	Specifically, a white-box source model is available for the target domain; thus, we term this task as \textit{white-box unsupervised domain adaptation} (WBUDA) to distinguish it from our investigated one in later paragraphs. In WBUDA, the adaptation could be achieved by fine-tuning the source model on unlabeled target data with well-designed objectives \cite{yang2020unsupervised, liang2020we}.

However, the white-box source model is not always given in practice. Most valuable models on cloud services (e.g., Google Cloud) are sealed as application programming interface (API), where only the input-output interface of a model is available and the model itself is kept as a black-box one. As stated in \cite{openai}, releasing an API instead of a white-box model could commercialize the technology, reduce model misuse and make the model use conveniently for the public; the white-box attacks \cite{thys2019fooling,tramer2017ensemble} may be also avoided. Due to all reasons mentioned above, white-box source models are probably unavailable in practice, which hinders the application of WBUDA methods.

In this work, we study a subtly different setting of source-free UDA, where only the API of the source model is accessible for the target domain. In other words, the source model itself is kept as a black-box one; thus, we term this task as \textit{black-box unsupervised domain adaptation} (B$^2$UDA). A few recent attempts \cite{yeh2021sofa,chidlovskii2016domain,morerio2020generative} have been made to tackle the B$^2$UDA problem, but achieving less satisfactory results.
In this work, we propose a simple yet effective algorithmic framework, termed Iterative Learning with Noisy Labels (IterLNL).
With black-box models as tools of noisy labeling, IterLNL conducts noisy labeling and LNL iteratively. 
Specifically, we first get model predictions of target data based on the black-box model and obtain their noisy labels as the category with the maximum prediction probability. We note that the label noise via the black-box model is highly unbalanced among categories (cf. Figure \ref{Fig:noise_visda}), which is significantly different from the simulated and balanced ones in LNL \cite{van2015learning,han2018co}; such unbalanced label noise hinders the application of state-of-the-art LNL methods \cite{jiang2018mentornet,han2018co}, inspiring the category-wise sampling strategy.
To facilitate the implementation of LNL in B$^2$UDA, we also estimate the noise rate from model predictions of unlabeled target data.
Experiments on benchmark datasets confirm the efficacy of our method. Notably, our IterLNL performs comparably with methods of traditional UDA setting where labeled source data are fully available. 	


\section{Related Work}

\noindent \textbf{Source Free UDA.} Traditional UDA \cite{dan,dann} assumes that labeled source data are available for the target domain. Due to increasing concerns for data privacy, source-free UDA \cite{liang2020we,Li_2020_CVPR,hou2020source,kim2020domain,kurmi2021domain,yang2020unsupervised} is highly appreciated as a new UDA setting, where only a source model is available for the target domain while labeled source data remain private. Source free UDA methods typically fine-tune the source model for the target domain with unlabeled target data \cite{liang2020we,kim2020domain,yang2020unsupervised,Li_2020_CVPR}. Specifically, Liang \emph{et al.} \cite{liang2020we} fine-tune the source model with pseudo-labeling and information maximization between target data and their predictions; a weight constraint is adopted in \cite{Li_2020_CVPR} to encourage similarity between the source model and adapted target model. Additionally, source data and source-style target data are respectively generated in \cite{kurmi2021domain} and \cite{hou2020source} using the statistics information stored in source model. The white-box source model is required in the methods above, but it may be unavailable due to the commercial and/or safety consideration \cite{openai}. To this end, we study a subtly different B$^2$UDA setting, where only the API of source model is accessible for the target domain; in other words, the source model itself is kept as a black-box one. We note that several attempts have been made on the B$^2$UDA problem recently. Based on pre-trained features, a denoising auto-encoder is used for prediction of target labels in \cite{chidlovskii2016domain} and an encoder-decoder framework is used in \cite{yeh2021sofa} where encoded target features are aligned to reference Gaussian distributions; however, both of the two methods obtain less satisfactory results on benchmark datasets. Morerio $\emph{et al.}$ \cite{morerio2020generative} first train a conditional Generative Adversarial Network (cGAN) with unlabeled target data and their source predictions, and then learn the target model with samples generated by cGAN; its performance is conditioned on the high-quality samples generated by cGAN, thus limiting its general usage in UDA tasks. In general, the B$^2$UDA problem is not well addressed yet. In the present work, we propose IterLNL and conduct thorough experiments on popular UDA benchmarks; results show that our proposed method works successfully for B$^2$UDA. 
	

\begin{figure*}
	\centering
	\includegraphics[width=0.8\textwidth]{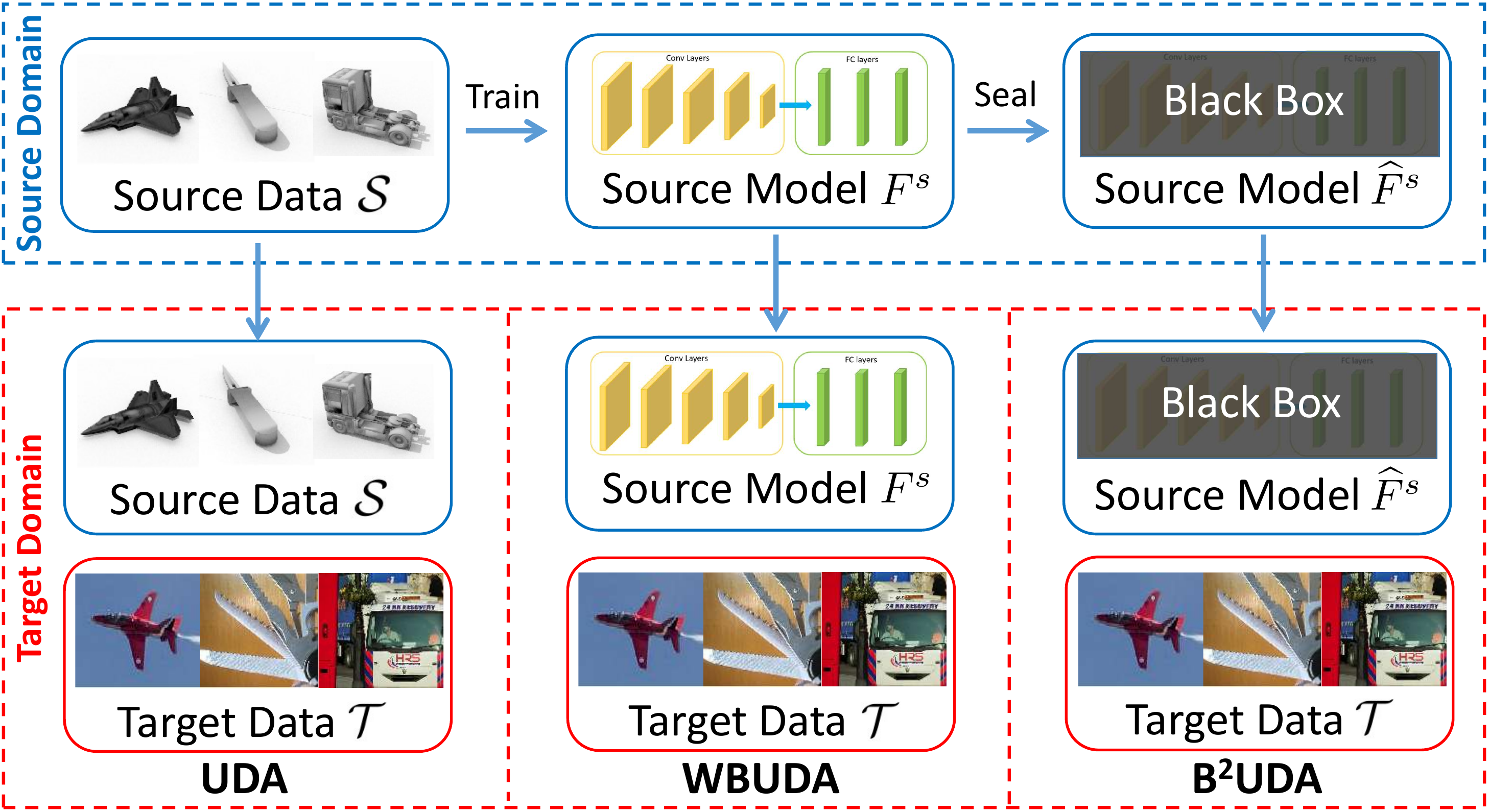}
	\caption{An illustration of different UDA settings. Source data and source model are respectively required in the traditional UDA and WBUDA settings. In contrast, B$^2$UDA requires a black-box access to the source model only, which is the least restrictive condition to apply domain adaptation to the unsupervised target data.}
	\label{Fig:comparison}
\end{figure*}
\begin{figure*}
	\centering
	\subfigure[Pair ($\epsilon$=0.45)] {
		\label{Fig:noise_pair}
		\includegraphics[width=0.23\textwidth]{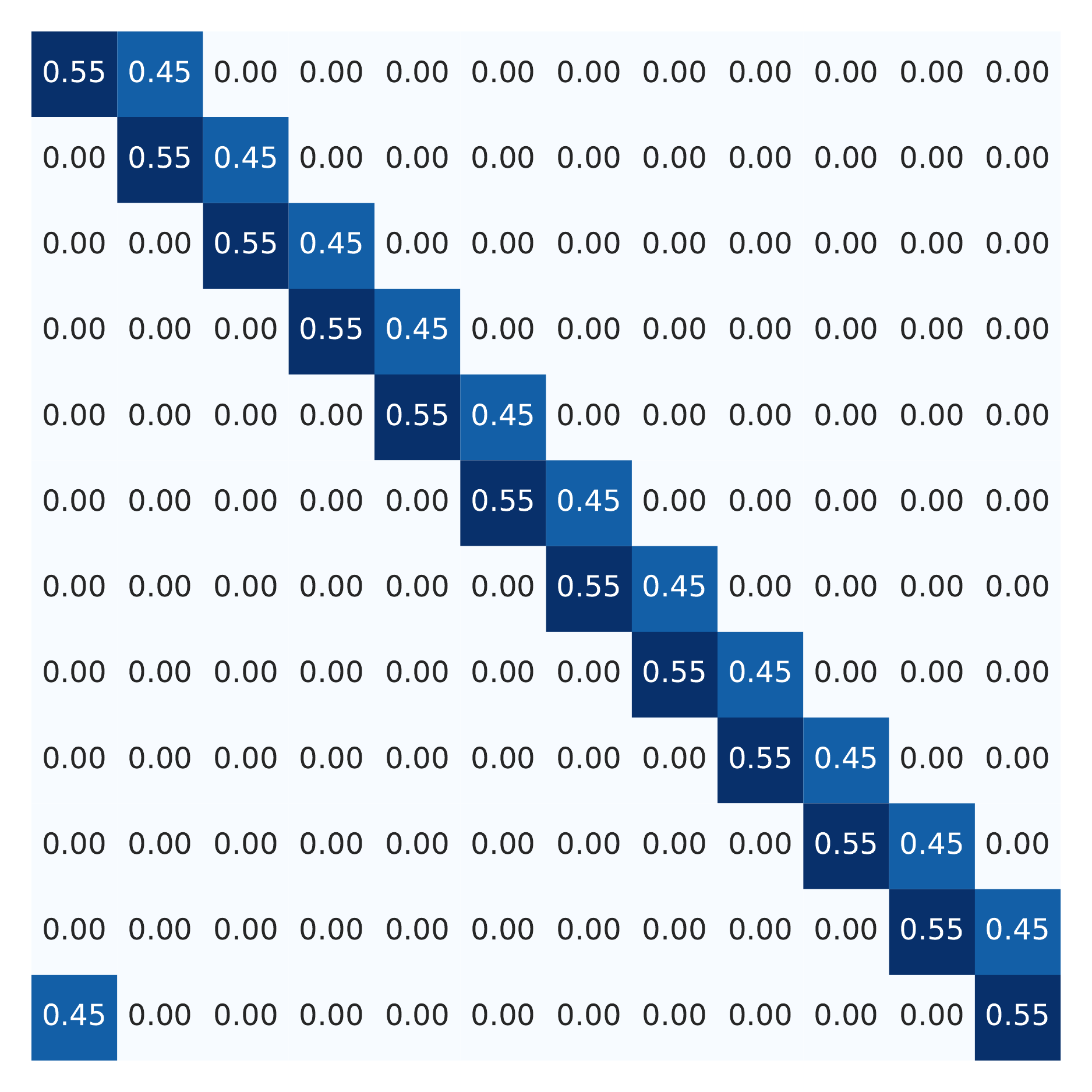}
	} \hfill
	\subfigure[Symmetry ($\epsilon$=0.5)] {
		\label{Fig:noise_sym}
		\includegraphics[width=0.23\textwidth]{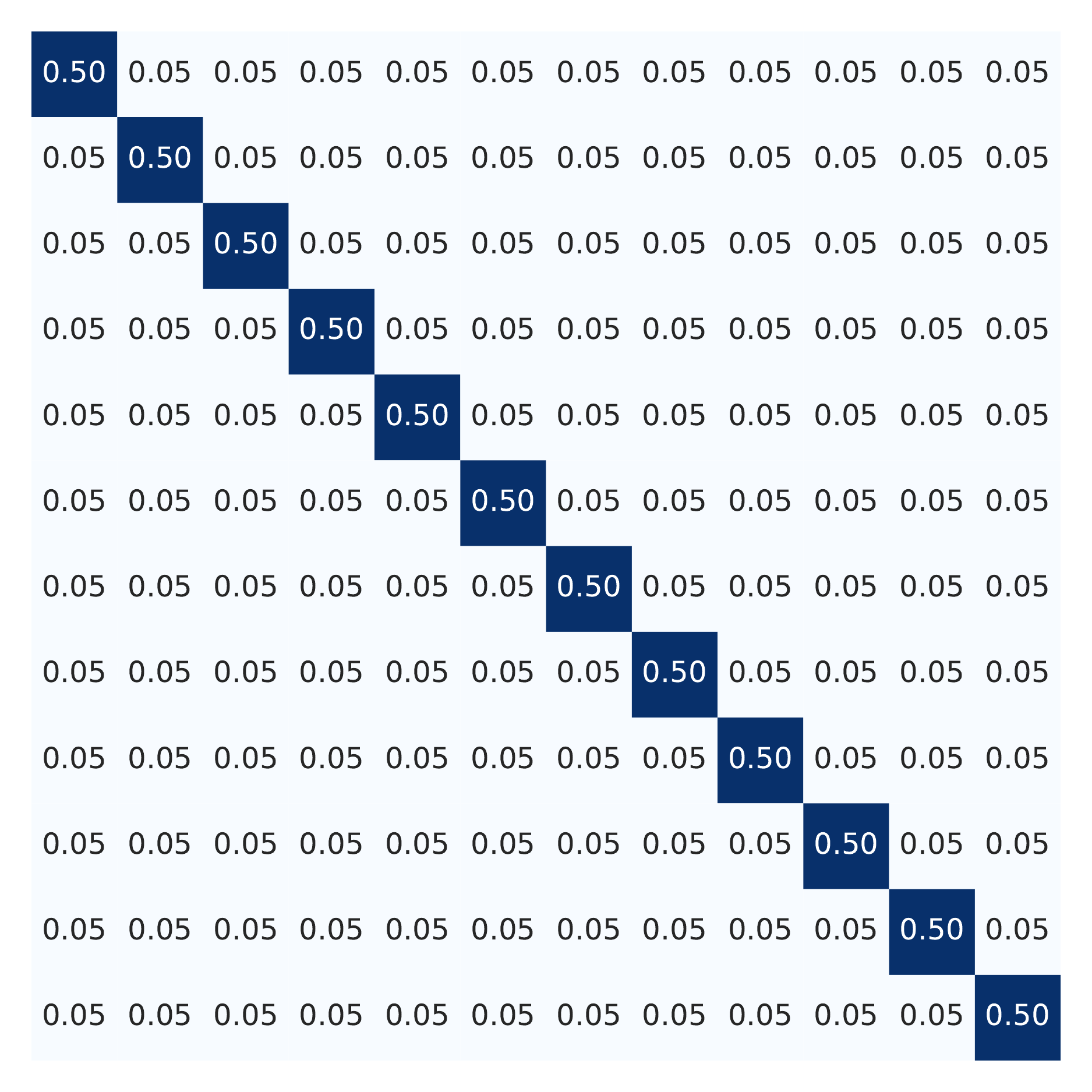}
	}
	\subfigure[VisDA-2017] {
	\label{Fig:noise_visda}
	\includegraphics[width=0.23\textwidth]{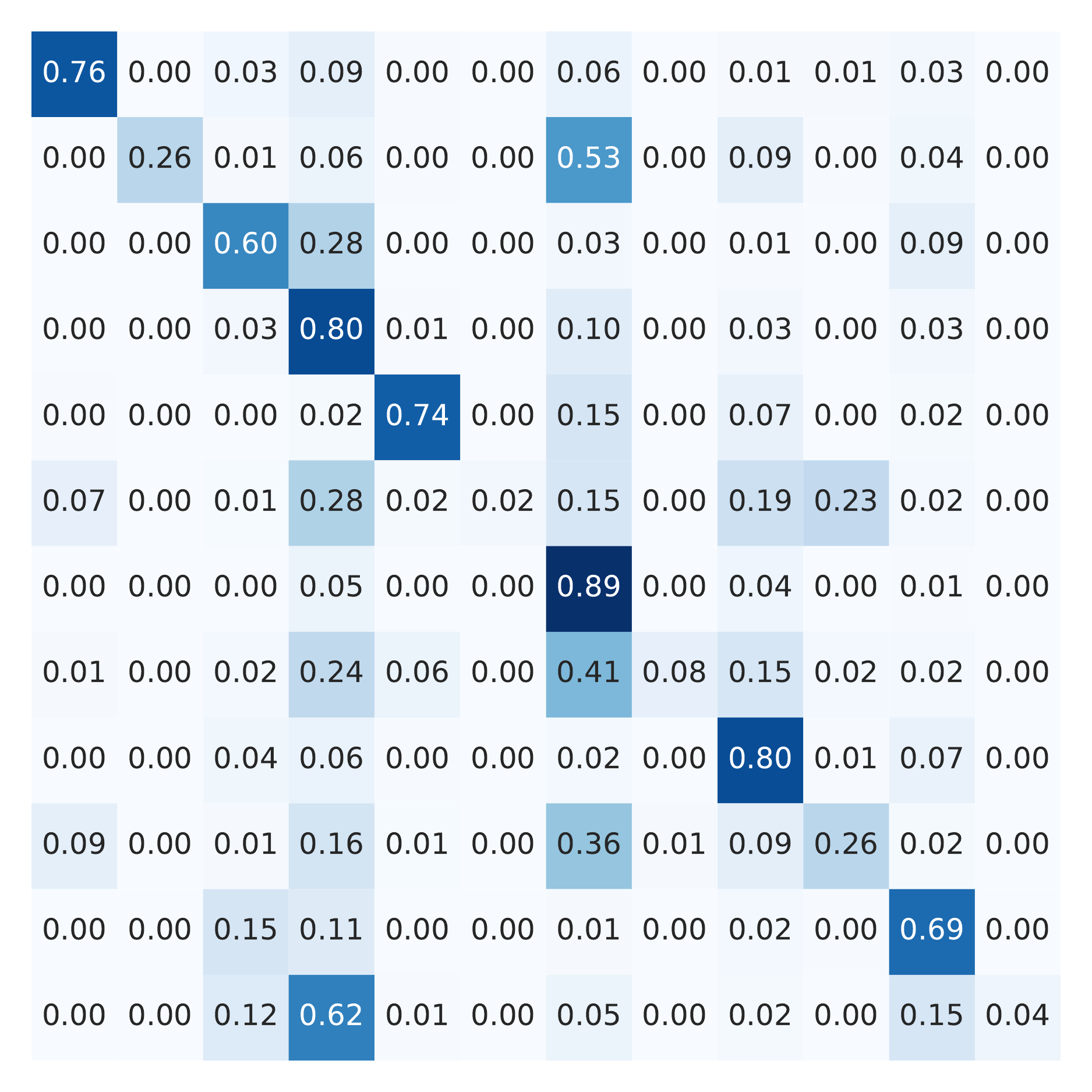}
}
	\subfigure[Rescale curve (\ref{Equ:rescale})] {
	\label{Fig:rescale_curve}
	\includegraphics[width=0.23\textwidth]{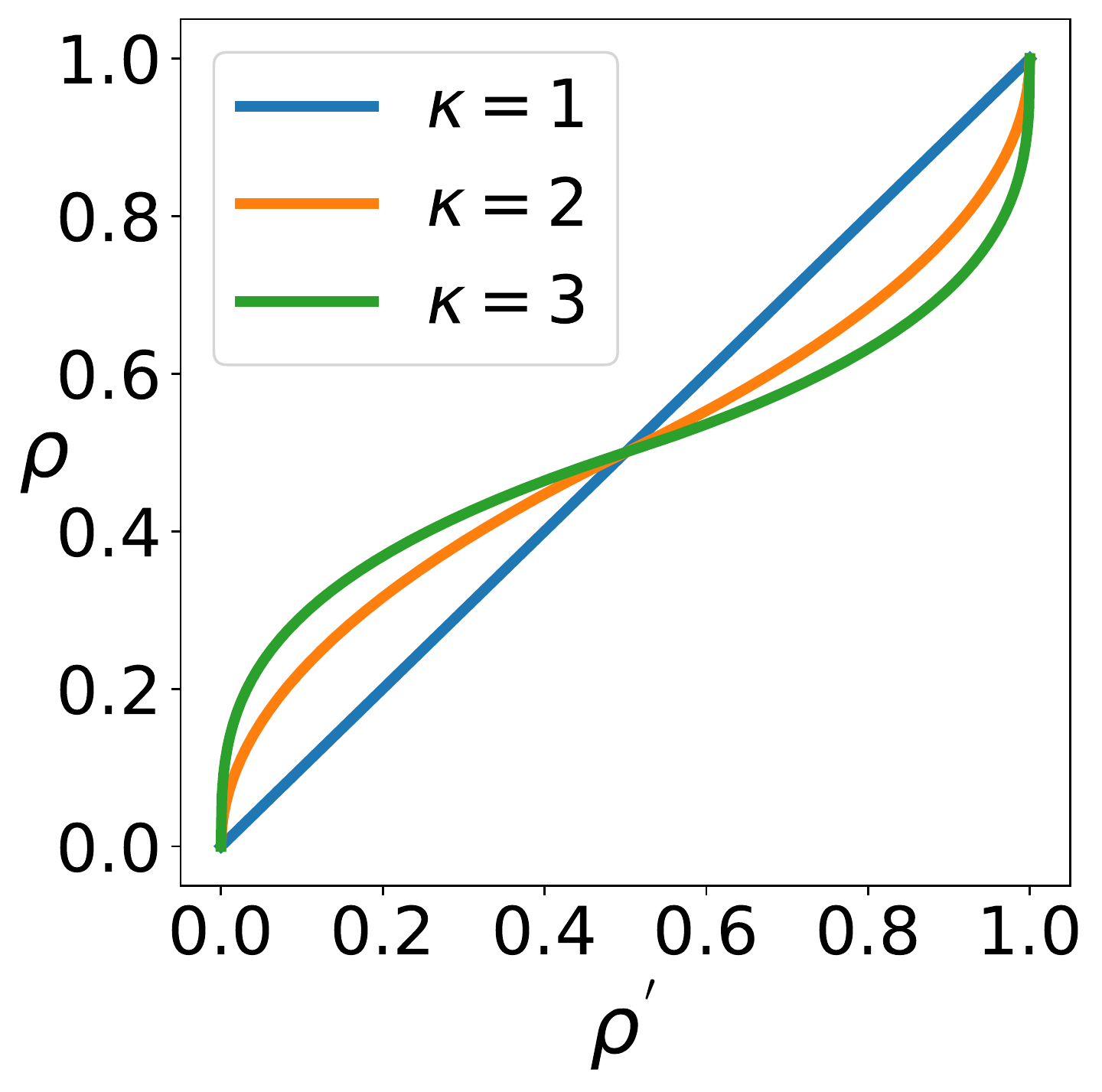}
}
	\caption{(a)-(c): Transition matrices \cite{patrini2017making, reed2014training} of different noise types, where the simulated (a) pair flipping \cite{han2018co} and (b) symmetry flipping \cite{van2015learning} are widely adopted in LNL works \cite{han2018co,yang2019searching}, and (c) presents the realistic noise matrix in the VisDA-2017 dataset based on the black-box source model $\widehat{F}^s$. The value in row $r$, column $c$ represents the probability with which samples of category $r$ are assigned with label $c$. In all figures, deeper colour indicates a larger value and all values are rounded to the level of $0.01$. (Zoom in to see the exact values.). The label noise in (c) VisDA-2017 is significantly unbalanced among categories, where the noise rates of $4$-th, $7$-th and $9$-th categories are less than 0.2 while the noise rates of $6$-th and $12$-th categories are more than 0.96.  (d) Illustration of the rescale curve (\ref{Equ:rescale}) with different $\kappa$. }  \label{Fig:noise_matrices}
\end{figure*}

\noindent \textbf{Learning with Noisy Labels (LNL).} LNL aims to learn models with noisy labeled data robustly. Seminal LNL methods include estimating noise transition matrix to transfer observed noisy labels to latent clean ones \cite{sukhbaatar2014training,van2017theory}, refining the objective function \cite{han2020sigua,zhang2018generalized}, and avoiding overfitting noisy labels with memorization effects of neuron networks \cite{jiang2018mentornet,han2018co}, as summarized in \cite{han2020survey}.
Although we also adopt LNL technical tools for the B$^2$UDA problem, the task divergence between LNL and B$^2$UDA raises new problems. Specifically, as predictions of black-box models are adopted as noisy labels, the noise rate is unknown and the label noise is unbalanced among categories in B$^2$UDA (cf. Figure \ref{Fig:noise_visda}), while simulated and balanced noisy labels with known noise rate are typically adopted in LNL \cite{van2015learning, han2018masking, han2018co}.
Besides, noisy labels are given and fixed in LNL, while we obtain noisy labels from predictions of black-box models and could update noisy labels via updating models, inspiring the iterative learning strategy.

\noindent \textbf{Hypothesis Transfer Learning (HTL).} HTL \cite{kuzborskij2013stability} utilizes source hypotheses (i.e., source models) to assist learning in the target domain. Different from the B$^2$UDA task, at least a small number of labeled target samples are demanded and source hypotheses are typically required to be white-box ones in HTL methods \cite{kuzborskij2013stability,yosinski2014transferable,ahmed2020camera}, hindering their applications in B$^2$UDA tasks.

\section{Problems and the proposed Method}
\label{Sec:Problem}
Given unlabeled target data $\mathcal{T}=\{ \x^t_i \}_{i=1}^{n^t}$ sampled from a distribution $\mathcal{Q}$, our problem of interest is to learn a model $F: \mathcal{X}^t \to [0,1]^K$ such that the empirical target risk $\frac{1}{n^t}\sum_{i=1}^{n^t} [\mathcal{L}(F(\x_i^t), y_i^t)]$ (or ideally, the expected risk $\mathbb{E}_{(\x^t, y^t) \in \mathcal{Q}} [\mathcal{L}(F(\x^t), y^t)]$) could be minimized, where $K$ is the category number, $\mathcal{L}$ is the loss function of the task, and $y^t_i \in \{1, \dots, K\}$, $i=1, \dots, n^t$, is the target label to be estimated. Depending on how much knowledge one may have from a source domain, the problem can fall in different established realms of unsupervised learning \cite{weber2000unsupervised}, unsupervised domain adaptation (UDA) \cite{dan,dann}, and source-free UDA \cite{yang2020unsupervised, liang2020we}. While the first one assumes no the source knowledge and is of machine learning foundations, in this work, we focus on different problem settings of UDA.

\vspace{0.1cm}
\noindent\textbf{Unsupervised Domain Adaptation.} Labeled source data $\mathcal{S} = \{ \x^s_i, y^s_i \}_{i=1}^{n^s}$ sampled from a distribution $\mathcal{P}$ are assumed to be available to help the learning with unlabeled target data $\mathcal{T}$ in UDA, leading to the following objective:
\begin{equation} \label{Equ:uda_objective}
F^t \Leftarrow \mathcal{O}_{\textrm{UDA}}(\mathcal{T}, \mathcal{S}),
\end{equation}
where $F^t$ is the expected target model. The most popular methods  \cite{dan,dann} for UDA (\ref{Equ:uda_objective}) are to learn domain invariant feature representations, then the classifier learned from labeled source data $\mathcal{S}$ could be safely applied to target data.

\vspace{0.1cm}
\noindent\textbf{White-box Unsupervised Domain Adaptation.} Source-free UDA \cite{Li_2020_CVPR, liang2020we} is proposed recently due to increasing concerns for data privacy. Indeed, we are in an era of cloud computing, and source and target data are usually captured and privately stored on different devices; it is thus risky to expose source data for transferable use to the target domain. Source-free UDA typically proposes to use a trained \textit{white-box source model} $F^s$, instead of the labeled source data $\mathcal{S}$, to accomplish the UDA objective, which is formalized as:
\begin{align} \label{Equ:sfda_wb_objective}
F^t \Leftarrow \mathcal{O}_{\textrm{WBUDA}}(\mathcal{T}, F^s),
\end{align}
where $F^s = \argmin_{F} \mathcal{L}_{task} (F, \mathcal{S})$ and the task loss $\mathcal{L}_{task}$ is typically instantiated as:
\begin{equation}
\mathcal{L}_{task} (F, \mathcal{S}) = \frac{1}{n^s}\sum_{i=1}^{n^s} -\log (F_{y_i^s}(\x_i^s)),
\end{equation}
where $F_k(\x)$ stands for the $k^{th}$ entry of $F(\x)$. Since white-box source model $F^s$ is required in (\ref{Equ:sfda_wb_objective}), we term it as \textit{white-box unsupervised domain adaptation} (WBUDA) to distinguish it from our investigated one in follow paragraphs.
In WBUDA, the target model is typically achieved by fine-tuning the white-box source model $F^s$ on unlabeled target data $\mathcal{T}$ using proper objectives, e.g., the pseudo labeling and information maximization losses in \cite{liang2020we}.

\noindent\textbf{Black-box Unsupervised Domain Adaptation.}
Although source data remain private in the WBUDA mentioned above \cite{Li_2020_CVPR, liang2020we}, the required white-box source model may be not available in practice. Most valuable models on cloud services (e.g., Google Cloud) are sealed as APIs, where only input-output interfaces are available. As stated in \cite{openai}, releasing an API instead of a white-box model could commercialize the technology, reduce model misuse and make the model use conveniently for the public; white-box attacks \cite{thys2019fooling,tramer2017ensemble} could also be avoided.

Considering above advantages of releasing APIs, we investigate a subtly different setting of source-free UDA, where only the API of a source model is accessible for the target domain and the source model itself is kept as a black-box one. We denote this task as \textit{black-box unsupervised domain adaptation} (B$^2$UDA), which is formulated as:
\begin{align} \label{Equ:sfda_bb_objective}
F^t \Leftarrow \mathcal{O}_{\textrm{B}^2\textrm{UDA}}(\mathcal{T}, \widehat{F}^s),
\end{align}
where $\widehat{F}^s$ is the API of $F^s$; specifically, we could get the output of $F^s(\x)$ with respect to any sample $\x$ via $\widehat{F}^s(\x)$ and the source model $F^s$ itself is kept as a black-box one. As there are many model APIs on cloud services, B$^2$UDA is a promising way to improve their adaptabilities, presenting broad practical values.

Labeled source data $\mathcal{S}$ and white-box source model $F^s$ are respectively required in UDA and WBUDA methods, impeding their applications in the B$^2$UDA task. 
To tackle the challenging B$^2$UDA, we propose an Iterative Learning with Noisy Labels (IterLNL) framework by conducting noisy labeling and LNL iteratively, which are introduced as follows.
\begin{figure*}[htbp]
	\centering
	\includegraphics[width=0.8\textwidth]{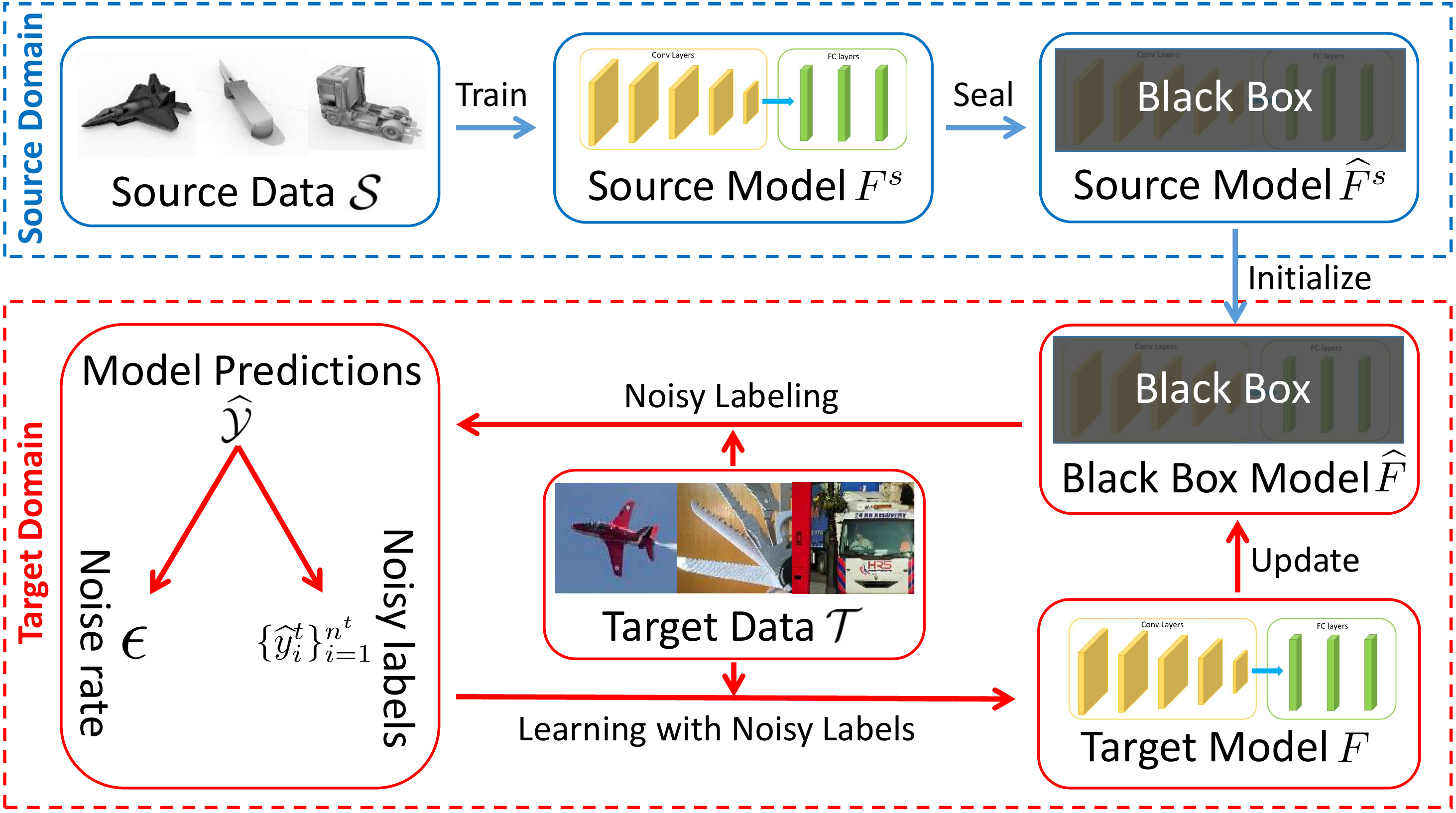}
	\caption{Framework of our proposed Iterative Learning with Noisy Labels (IterLNL), where we conduct noisy labeling and LNL iteratively. Note that we introduce noisy labels based on predictions of the black-box model.}
	\label{Fig:framework}
\end{figure*}

\subsection{Noisy Labeling} \label{Sec:noisy_labeling}
Given a black-box model $\widehat{F}$ (e.g., the black-box source model $\widehat{F}^s$) in B$^2$UDA, we could get label predictions of target data $\{ \x_i^t\}_{i=1}^{n^t}$ with $\widehat{F}$ as:
	\begin{equation} \label{Equ:noisy_labeling}
	\widehat{\mathcal{Y}} = \{\widehat{F}(\x_i^t)\}_{i=1}^{n^t}.
	\end{equation}
	The corresponding pseudo label of target sample $\x_i^t$  is defined as $\widehat{y}_i^t = \argmax_k\widehat{F}_k(\x_i^t)$. The pseudo labels $\{\widehat{y}_i^t\}_{i=1}^{n^t}$ could be highly noisy due to the divergence across source and target domains. Furthermore, we emphasize that the label noise in $\{\widehat{y}_i^t\}_{i=1}^{n^t}$ could be significantly unbalanced among categories; for example, the noise rate could be extremely high for some categories and extremely low for the others, as illustrated in Figure \ref{Fig:noise_visda}.
	Such unbalanced label noise via domain shift is substantially different from the simulated ones \cite{van2015learning,han2018co} in many LNL works, as compared in Figure \ref{Fig:noise_matrices}. In addition, the noise rate, which is usually required in LNL algorithms, is unknown in B$^2$UDA, while it is usually assumed to be given in LNL \cite{van2015learning, han2018co}. In the next paragraph, we propose strategies to estimate the noise rate and tackle unbalanced label noise, which support the successful target model learning with noisy labels.

\subsection{Learning with Noisy Labels} \label{Sec:LNL}
Give noisy target label predictions $\widehat{\mathcal{Y}}$ in Section \ref{Sec:noisy_labeling}, we resort to LNL to learn the target model.
State-of-the-art LNL methods \cite{jiang2018mentornet,han2018co,hendrycks2016baseline} usually combat noisy labels by selecting `clean' samples from each mini-batch for training, which is achieved by utilizing the memorization effects of neuron networks \cite{arpit2017closer}. Before going into detail, we denote $R(n)$ as the percentage of instances selected for training in the mini-batch of $n$-th iteration. LNL methods \cite{jiang2018mentornet,han2018co,hendrycks2016baseline} typically keep more instances in the mini-batch (i.e., $R(n)$ is large) at the beginning, and then gradually drop noisy samples  (i.e., $R(n)$ becomes smaller) as training proceeds; by using more training instances at the beginning, a relatively reliable model could be achieved since deep models learn clean and easy patterns at the beginning \cite{arpit2017closer}; with the reliable model, noisy instances could be filtered out by gradually dropping instances with larger losses.

We also adopt the aforementioned LNL strategy in our method since it presents high robustness even with extremely noisy labels \cite{han2018co}. In LNL \cite{han2018co,yang2019searching}, the selecting percentage $R(n)$ is depending on the noise rate, which is either assumed to be known in advance \cite{han2018co} or estimated with few labeled clean data \cite{yu2018efficient,ramaswamy2016mixture}.
However, realistic noisy labels are introduced by domain shift in B$^2$UDA, where the unavailable of labeled target data and unknown noise rate impede the design of $R(n)$.

To this end, we propose a simple yet efficient strategy to estimate noise rate. We first follow \cite{han2018co} to define $R(n)$ as:
\begin{equation} \label{Equ:sample_strategy}
R(n)=1- \min{(\frac{n}{n_k} \epsilon, \epsilon)},
\end{equation}
where $n_k$ is a hyperparameter and $\epsilon$ is the noise rate.

\noindent\textbf{Noise rate Estimation.}  We first present the empirical noise rate $\epsilon$ as:
\begin{equation} \label{Equ:noise_level_empirical}
\epsilon = 1- \frac{1}{n^t} \sum_{i=1}^{n^t} \mathcal{I}[\x^t_i, \widehat{y}_i^t ],
\end{equation}
where
$\mathcal{I}[\x^t_i, \widehat{y}_i^t ] \in \{0,1\}$ is a binary indicator;  $\mathcal{I}[\x^t_i, \widehat{y}_i^t ] =1$ if $\widehat{y}_i^t$ is the correct label of $\x^t_i$ and 0 otherwise. It is obvious that the empirical noise rate $\epsilon$ (\ref{Equ:noise_level_empirical}) is close correlated to the classification accuracy of the black-box model $\widehat{F}$. In the meantime, there is a correlation between the classification accuracy and maximum prediction probability, as observed in \cite{lee2013pseudo, zhou2020time, hendrycks2016baseline}. Although the prediction probability may be overconfident and misleading viewed in isolation, the probability statistics is often sufficient to reflect on the overall classification accuracy \cite{hendrycks2016baseline,guo2017calibration}, and also the noise rate $\epsilon$.

To estimate the noise rate $\epsilon$, we calculate the proportion of target data $\mathcal{T}$ with high prediction probability as:
\begin{equation} \label{Equ:proportion_high_confidence}
\rho' = \frac{1}{n^t} \sum_{i=1}^{n^t} \mathbb{I}[ \max(\widehat{F}(\x_i^t)) > \gamma ],
\end{equation}
where $\gamma \in [0,1]$ is the threshold and $\mathbb{I} [var] = \begin{cases}
1, & var = True\\
0, & Otherwise
\end{cases}$. One may opt for approximating noise rate $\epsilon$ with $1-\rho'$. However, considering that deep models are prone to make over-confident predictions \cite{guo2017calibration}, we rescale $\rho'\in[0,1] \to \rho \in[0,1]$ as:
\begin{equation} \label{Equ:rescale}
 \rho = \begin{cases}
 0.5(2\rho')^{1 / \kappa}, & \rho' < 0.5\\
 -0.5(2-2\rho')^{1 / \kappa} + 1, & Otherwise.
 \end{cases}
\end{equation}
Although the equation (\ref{Equ:rescale}) seems complicated, there is only one hyperparameter $\kappa$ controlling the curve degree and it degenerates to $\rho =\rho'$ if $\kappa=1$, as illustrated in Figure \ref{Fig:rescale_curve}. The design of (\ref{Equ:rescale}) is further investigated in Section \ref{Sec:ablation}. We finally approximate noise rate $\epsilon$ as:
\begin{equation} \label{Equ:noise_level}
\epsilon = 1 - \rho.
\end{equation}
Although the estimated noise rate $\epsilon$ (\ref{Equ:noise_level}) is not precise, we find that our method is robust to the noise rate $\epsilon$; such an estimation of noise rate works well and achieves good results close to that using the grounding truth noise rate, as presented in Section \ref{Sec:ablation}.

\noindent \textbf{Category-wise Sampling.} Given the estimated noise rate $\epsilon$ (\ref{Equ:noise_level}), we could conduct LNL by selecting $R(n)$ (\ref{Equ:sample_strategy}) percent samples with smaller loss for training in the mini-batch of $n$-th iteration. However, as we state in Section \ref{Sec:noisy_labeling}, the label noise is unbalanced among categories in B$^2$UDA (cf. Figure \ref{Fig:noise_visda}); thus samples in categories with higher noise rate are prone to present larger loss and be rejected for training, leading to worse results for these categories, as presented in Table \ref{Table:ablation_visda}.

To this end, we propose to individually sample the $R(n)$ (\ref{Equ:sample_strategy}) percent samples with smaller loss for each category. Technically, we introduce a probability queue buffer $\u_k \in [0,1]^{h}$ for category $k\in[1,\dots,K]$, where $\u_k$ is initialized as a vector filled with positive infinity values and $h$ is the buffer length. For any instance $\x^t$ in the $n$-th iteration, we obtain its corresponding noisy label $\widehat{y}^t=\argmax_k\widehat{F}_k(\x^t)$ and loss $\mathcal{L}(F(\x^t), \widehat{y}^t) = -\log(F_{\widehat{y}^t}(\x^t))$, where $F$ is the current model in learning. We propose an indicator $\I(\mathcal{L}(F(\x^t), \widehat{y}^t), \u_{\widehat{y}^t}, n)$ to decide whether $\x^t$ should be adopted in training, which is formulated as:
\begin{equation} \label{Equ:indicator}
\I(\mathcal{L}(F(\x^t), \widehat{y}^t), \u_{\widehat{y}^t}, n) = \begin{cases}
1, & \mathcal{L}(F(\x^t),\widehat{y}) \leq L_{R(n)}(\u_{\widehat{y}^t})  \\
0, & Otherwise,
\end{cases}
\end{equation}
where $L_{R(n)} (\u_{\widehat{y}^t})$ is the $R(n)$-th largest value in $\u_{\widehat{y}^t}$. We utilize the loss $\mathcal{L}(F(\x^t), \widehat{y}^t)$ to update current model $F$ if $\I(\mathcal{L}(F(\x^t), \widehat{y}^t), \u_{\widehat{y}^t}, n)=1$ and drop it otherwise.

We also update the queue buffers $\{\u_k\}_{k=1}^{K}$ with all samples in the $n$-th iteration. Specifically, given an instance $\x^t$ and its corresponding noisy label $\widehat{y}^t$ and loss $\mathcal{L}(F(\x^t), \widehat{y}^t)$ defined above, we push $\mathcal{L}(F(\x^t), \widehat{y}^t)$ into the queue buffer $\u_{\widehat{y}^t}$ and pop the oldest value from $\u_{\widehat{y}^t}$ simultaneously. In this way, we adopt samples with the $R(n)$ percent smallest losses in each category for training in $n$-th iteration.

\vspace{0.1cm}
\noindent \textbf{Remarks.}
Note that our method is fundamentally different from existing DA methods based on pseudo labels \cite{jan,zou2018unsupervised}. Although we also adopt the categories with maximum prediction probability on the black-box model as noisy labels, we do not use them directly as accurate ones. We only use samples with noisy labels to update the model if these samples present small losses based on the current model; in other words, we treat noisy labels as accurate ones only if they are consistent with the current model's predictions. Moreover, source data are not involved in model training, avoiding the misleading of source data on target predictions.

\subsection{Iternative Learning Strategy} \label{Sec:iterative_learning}

\begin{figure}
	\centering
	\subfigure[Acc. in iterative process] {
		\label{Fig:acc_iterative}
		\includegraphics[width=0.225\textwidth]{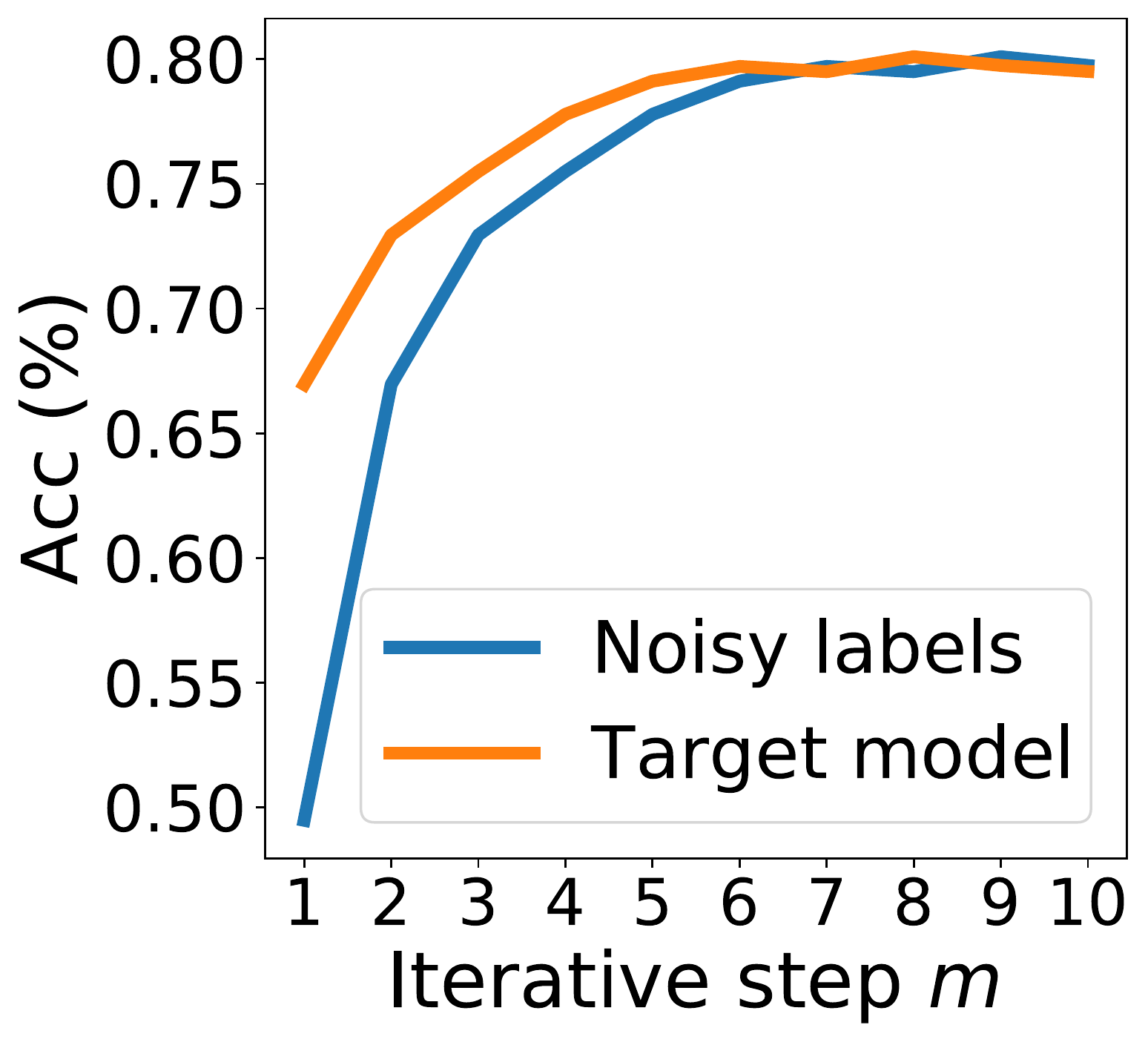}
	} \hfill
	\subfigure[Noise rate $\epsilon$] {
		\label{Fig:epsilon}
		\includegraphics[width=0.225\textwidth]{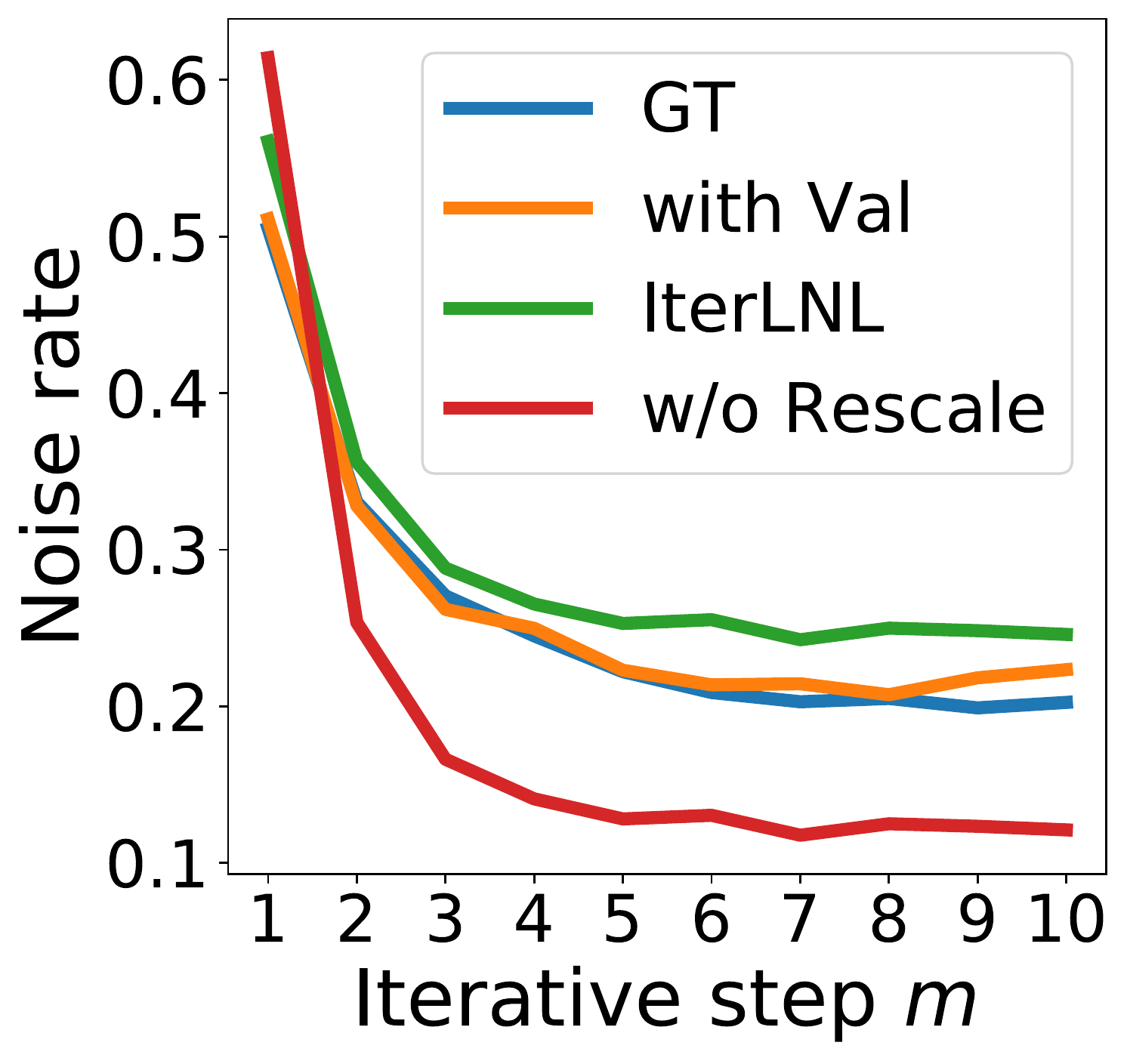}
	}
	\caption{Illustration of (a) the accuracy of noisy labels $\{ \widehat{y}_i^t\}_{i=1}^{n^t}$ and target model $F$ via LNL, and (b) the estimated noise rate in different iterative steps. `GT', `with Val', `IterLNL' and `w/o Rescale' indicate the noise rate calculated by all labeled target data (\ref{Equ:noise_level_empirical}), labeled validation data, our strategy (\ref{Equ:noise_level}), and $1-\rho'$, respectively.
		Note that the noise rate estimation and accuracy of noisy labels with $m=1$ are based on predictions of the black-box source model $\widehat{F}^s$.  }
\end{figure}

With the noisy labeling in Section \ref{Sec:noisy_labeling} and LNL in Section \ref{Sec:LNL}, we could get a more reliable target model over the original black-box one (i.e., the one introduces noisy labels), as illustrated in Figure \ref{Fig:acc_iterative}. In other words, the achieved target model could produce improved noisy labels over the original noisy labels. It is a natural idea to conduct noisy labeling with the achieved target model (or the black-box counterpart of the achieved target model), and conduct LNL on the new noisy labels again. Finally, we define our algorithmic framework by conducting noisy labeling and LNL iteratively, leading to the Iterative Learning with Noisy Labels (IterLNL). We summarize IterLNL in Algorithm \ref{Alg:IterLNL} and illustrate its framework in Figure \ref{Fig:framework}.

\vspace{0.1cm}
\noindent \textbf{Remarks.}
The black-box source model and noisy labels are introduced respectively in B$^2$UDA and LNL tasks to help the learning with unlabeled data, making the two tasks fundamentally different. Although we utilize the black-box source model to assign noisy labels and learn the target model in a LNL manner, there are three differences. First, the noise rate is unknown and to be estimated in B$^2$UDA and no labeled target data are provided for the noise rate estimation. Second, the label noise is quite unbalanced among categories in B$^2$UDA. Third, different from the given and fixed noisy labels in LNL, the noisy labels in B$^2$UDA are introduced by black-box models and could be updated as the model update, inspiring the iterative learning strategy.



\begin{algorithm}
	\caption{Iterative Learning with Noisy Labels.}
	\label{Alg:IterLNL}
	\begin{algorithmic}[1]
		\Require Black-box source model $\widehat{F}^s$, target data $\mathcal{T}$
		\Ensure Target model $F$
		\State Initialize $\widehat{F}$ with $\widehat{F}^s$
		\For {$m=1$ to $M$} \algorithmiccomment{For each iterative step}
		\State Acquire noisy labels $\widehat{\mathcal{Y}}$ with $\mathcal{T}$ and $\widehat{F}$ using (\ref{Equ:noisy_labeling})
		\State Estimate noise rate $\epsilon$ with (\ref{Equ:proportion_high_confidence}), (\ref{Equ:rescale}) and (\ref{Equ:noise_level})
		\State Initialize target model $F$ and buffers $\{ \u_k\}_{k=1}^K$
		\For {$n=1$ to $N$} \algorithmiccomment{For each iteration}
		 	\State Acquire  $R(n)$ with (\ref{Equ:sample_strategy})
		 	\State Update model $F$ using data selected with (\ref{Equ:indicator})
		 	\State Update buffers $\{ \u_k\}_{k=1}^K$
		\EndFor
		\State Update $\widehat{F}$ as the API (i.e., black-box model) of $F$
		\EndFor
	\end{algorithmic}
\end{algorithm}

\begin{table*}[h!]
	\centering
	\begin{tabular}{l|p{0.70cm}<{\centering}p{0.70cm}<{\centering}p{0.70cm}<{\centering}p{0.70cm}<{\centering}p{0.70cm}<{\centering}p{0.70cm}<{\centering}p{0.70cm}<{\centering}p{0.70cm}<{\centering}p{0.70cm}<{\centering}p{0.70cm}<{\centering}p{0.70cm}<{\centering}p{0.70cm}<{\centering}|c}
		\hline
		Methods  & plane & bcycl & bus & car & horse & knife & mcycl & person & plant & sktbrd & train & truck & Avg \\
		\hline
		Source Model \cite{dalib} & 76.4 & 26.4 & 60.2 & 80.2 & 73.5 & 2.3 & 89.4 & 8.2 & 79.7 & 25.6 & 69.4 & 3.8 & 49.6 \\
		IterLNL (w/o Iter) & 91.4 & 68.9 & 69.6 & 89.9 & 88.0 & 25.1 & \textbf{92.8} & 26.1 & 93.1 & 59.1 & 83.3 & 16.3 & 67.0 \\
		IterLNL (w/o CateS) & \textbf{96.7} & \textbf{88.3} & \textbf{89.0} & \textbf{94.1} & \textbf{96.7} & 0.0 & 89.4 & 48.4 & \textbf{96.0} & 93.3 & 84.8 & 0.0 & 73.1 \\
		IterLNL (w/o Rescale) & 91.7 & 84.8 & 86.6 & 88.7 & 79.4 & 77.9 & \textbf{92.8} & 46.3 & 90.1 & 89.1 & \textbf{88.2} & 40.5 & 79.7 \\
		\hline
		IterLNL & 77.0 & 84.6 & 85.1 & 92.0 & 92.1 & 74.1 & 92.6 & \textbf{49.1} & 89.1 & 91.7 & 84.5 & 49.4 & 80.1 \\
		\hline
		\hline
		IterLNL (with Val) & 89.0 & 79.5 & 84.3 & 81.0 & 87.7 & \textbf{88.1} & 92.5 & 38.7 & 87.1 & \textbf{96.9} & 78.8 & \textbf{67.0} & \textbf{80.9} \\
		\hline
	\end{tabular}
	\caption{Ablation study on VisDA-2017 dataset, where all experiments are based on a ResNet-101 model. Please refer to the main text for definitions.}
	\label{Table:ablation_visda}	
\end{table*}

\section{Experiment} \label{Sec:Experiment}
\noindent \textbf{Office-31} \cite{saenko2010adapting} is the most popular benchmark dataset for UDA. There are $4,110$ samples shared by three domains: amazon (A), webcam (W), and dslr (D).  \textbf{VisDA-2017} \cite{peng2017visda} aims to transfer knowledge from synthetic images to real-world ones, which is a challenging task with significant domain divergence. There are $152K$ synthetic images and $55K$ real-world images shared by $12$ classes.
Datasets of MNIST \cite{lecun1998gradient}, Street View House Numbers (SVHN) \cite{netzer2011reading}, and USPS \cite{hull1994database} constitute the \textbf{Digits} task, which includes $10$ classes. There are $50,000$ training samples, $10,000$ validation samples and $10,000$ test samples in the MNIST dataset (M), where all images are black-and-white handwritten digits; the SVHN dataset (S) contains $73,257$ training and $26,032$ test images with colored backgrounds; the USPS dataset (U) contains $7,291$ training and $2,007$ test images with black backgrounds.

\noindent \textbf{Implementation Details.}
For experiments on datasets of Office-31 and VisDA-2017, we employ the pre-trained ResNet model \cite{resnet} as the backbone and replace the last fully connected (FC) layer with a task-specific FC classifier following \cite{dan,dann, liang2020we}. We introduce the source model $F^s$ by fine-tuning the constructed model on source data following \cite{dalib} and then seal $F^s$ as the black-box $\widehat{F}^s$, i.e., only the input-output interface of $F^s$ is available. For experiments on Digits dataset, we follow \cite{mcd} to introduce the source model $F^s$ with convolutional layers and FC layers. Following \cite{dann}, we utilize the SGD optimizer and adopt the learning rate strategy as $\eta_n = \frac{\eta_0}{(1+10\zeta)^{0.75}}$, where $\eta_0=0.01$ and $\zeta$ is the process of training iterations linearly changing from $0$ to $1$.
We set the batch size as $64$, $\kappa=2$ in (\ref{Equ:rescale}), $\gamma=0.9$ in (\ref{Equ:proportion_high_confidence}), buffer length $h=100$, and $n_k=0.5N$ ($N$ is the total number of training iterations defined in Algorithm \ref{Alg:IterLNL}) in (\ref{Equ:sample_strategy}) for all experiments; these hyperparameters are analyzed in Section \ref{Sec:analysis}.

\subsection{Ablation Study}
\label{Sec:ablation}

We introduce several variants of IterLNL to investigate the individual components in IterLNL.  Specifically, following \cite{han2018co}, we replace the category-wise sampling (\ref{Equ:indicator}) by simply using the $R(n)$ percent samples with smaller losses in the $n$-th iteration for model learning, leading to `IterLNL (w/o CateS)'. We also present the results of IterLNL by conducting noisy labeling and learning with noisy labels only once (i.e., setting $M=1$ in Algorithm \ref{Alg:IterLNL}), resulting in `IterLNL (w/o Iter)'. To investigate the noise rate estimation strategy, we remove the rescale strategy (\ref{Equ:rescale}) and set $\epsilon = 1-\rho'$, leading to `IterLNL (w/o Rescale)'; we also introduce a labeled target validation set by randomly selecting $30$ samples per class (less than $1$\% of the entire target data in VisDA-2017); we estimate the noise rate $\epsilon$ by calculating the classification accuracy $\alpha$ of $\widehat{F}$ on the validation set and approximating the noise rate $\epsilon$ as $1 - \alpha$, which is termed as `IterLNL (with Val)'.

As illustrated in Table \ref{Table:ablation_visda}, IterLNL improves over the IterLNL (w/o CateS) and IterLNL (w/o Iter) significantly, justifying the efficacy of category-wise sampling (\ref{Equ:indicator}) and iterative learning strategy. Specifically, without the category-wise sampling (\ref{Equ:indicator}), the accuracy of samples in the knife and truck categories drops to zero due to the initial high noise rate (i.e., low accuracy) in Source Model. We also intuitively visualize the accuracy improvement of model $F$ in the iterative learning process (i.e., with different $m \in [1, M]$). As illustrated in Figure \ref{Fig:acc_iterative}, the accuracy of $F$ via LNL indeed improves over that of initial noisy labels in the beginning; the improvement is gradually reduced as the iterative step $m$ increases, leading to the final convergence.
Additionally, IterLNL improves over IterLNL (w/o Rescale) and approximate to results of IterLNL (with Val); this is intuitively explained in Figure \ref{Fig:epsilon}, where IterLNL (with Val) presents the most accurate noise rate estimation, and IterLNL achieves more accurate estimation than IterLNL (w/o Rescale) in most cases. The source model (i.e., $m$=1) tends to make un-confident predictions on target data due to the domain divergence while target models (i.e., $m$>1) are prone to make over-confident predictions, motivating us to introduce the rescale curve (\ref{Equ:rescale}) in Figure \ref{Fig:rescale_curve}.

\begin{table}
	\centering
	\begin{tabular}{l|l|p{1.2cm}<{\centering}p{1.2cm}<{\centering}p{1.2cm}<{\centering}}
		\hline
		Settings & Methods & U$\to$M & S$\to$M & M$\to$U \\
		\hline
		\multirow{4}{*}{B$^2$UDA} & Source Model  & 82.0 & 69.4 & 79.4 \\
		& sMDA \cite{chidlovskii2016domain} & 83.4 & 69.9 & 81.2\\
		& PLR \cite{morerio2020generative}    & 91.8 & 97.3 & 89.3 \\
		& IterLNL    & \textbf{96.7}$\pm$0.3 & \textbf{98.0}$\pm$0.7 & \textbf{97.4}$\pm$0.1 \\
		\hline
		\hline
		\multirow{3}{*}{WBUDA} & SDDA \cite{kurmi2021domain} & -- & 75.5 & 89.9 \\
		&3C-GAN \cite{Li_2020_CVPR}  & 99.3$\pm$0.1 & 99.4$\pm$0.1 & 97.3$\pm$0.2  \\
		& SHOT \cite{liang2020we} & 98.4 & 98.9 & 98.0 \\
		\hline
		\hline
		\multirow{4}{*}{UDA} & DANN \cite{dann} & 86.3$\pm$0.3 & 85.5$\pm$0.4 & 84.9$\pm$0.6 \\
		&MCD \cite{mcd} & -- & 96.2$\pm$0.4 & 96.5$\pm$0.3 \\
		& CDAN \cite{cada} & 98.0 & 89.2 & 95.6 \\
		& RWOT \cite{xu2020reliable} & 97.5$\pm$0.2 & 98.8$\pm$0.1 & 98.5$\pm$0.2 \\
		\hline
	\end{tabular}
	\caption{Results on Digits dataset.}
	\label{Table:digit}
\end{table}

\begin{table*}[h!]
	\centering
	\begin{tabular}{l|l|cccccc|c}
		\hline
		Tasks & Method & A$\to$D & A$\to$W & D$\to$A & D$\to$W & W$\to$A & W$\to$D & Avg \\
		\hline
		\multirow{3}{*}{B$^2$UDA}&Source Model \cite{dalib} & 79.7 & 78.1 & 64.9 & 96.0 & 65.4 & 99.2 & 80.1 \\
		& sMDA \cite{chidlovskii2016domain} & 80.5 & 79.3 & 65.7 & 96.3 & 67.3 & 99.2 & 81.4 \\
		& IterLNL & \textbf{91.3}$\pm$0.1 & \textbf{89.8}$\pm$0.7 & \textbf{74.3}$\pm$0.8 & \textbf{98.7}$\pm$0.2 & \textbf{73.7}$\pm$0.1 & \textbf{99.3}$\pm$0.3 & \textbf{87.9}  \\
		\hline
		\hline
		\multirow{3}{*}{WBUDA}		& SDDA \cite{kurmi2021domain} & 85.3 & 82.5 &  66.4 & 99.0 & 67.7 & 99.8 & 83.5 \\
		&SHOT \cite{liang2020we} & 94.0 & 90.1 & 74.7 & 98.4 & 74.3 & 99.9 & 88.6 \\
		&3C-GAN \cite{Li_2020_CVPR} & 92.7$\pm$0.4 & 93.7$\pm$0.2 & 75.3$\pm$0.5 & 98.5$\pm$0.1 & 77.8$\pm$0.1 & 99.8$\pm$0.2 & 89.6 \\
		\hline
		\hline
		\multirow{4}{*}{UDA} & DANN \cite{dann} & 79.7$\pm$0.4 & 82.0$\pm$0.4 & 68.2$\pm$0.4 & 96.9$\pm$0.2 & 67.4$\pm$0.5 & 99.1$\pm$0.1 & 82.2 \\
		& CDAN \cite{cada} & 92.9$\pm$0.2 & 94.1$\pm$0.1 & 71.0$\pm$0.3 & 98.6$\pm$0.1 & 69.3$\pm$0.3 & 100.0$\pm$.0 & 87.7 \\
		& SymNets \cite{symnets} & 93.9$\pm$0.5 & 90.8$\pm$0.1 & 74.6$\pm$0.6 & 98.8$\pm$0.3 & 72.5$\pm$0.5 & 100.0$\pm$.0 & 88.4  \\
		& RWOT \cite{xu2020reliable} & 94.5$\pm$0.2 & 95.1$\pm$0.2 & 77.5$\pm$0.1 & 99.5$\pm$0.2 & 77.9$\pm$0.3 &  100.0$\pm$.0 & 90.8 \\
		\hline
	\end{tabular}
	\caption{Results on Office31 dataset, where all methods are based on a ResNet-50 model. }
	\label{Table:office31}
\end{table*}

\begin{table*}[h!]
	\centering
	\begin{tabular}{l|l|p{0.62cm}<{\centering}p{0.62cm}<{\centering}p{0.62cm}<{\centering}p{0.62cm}<{\centering}p{0.62cm}<{\centering}p{0.62cm}<{\centering}p{0.62cm}<{\centering}p{0.62cm}<{\centering}p{0.62cm}<{\centering}p{0.62cm}<{\centering}p{0.62cm}<{\centering}p{0.62cm}<{\centering}|c}		
		\hline
		\rotatebox{0}{Tasks} & \rotatebox{0}{Methods} & \rotatebox{60}{plane} & \rotatebox{60}{bcycl} & \rotatebox{60}{bus} & \rotatebox{60}{car} & \rotatebox{60}{horse} & \rotatebox{60}{knife} & \rotatebox{60}{mcycl} & \rotatebox{60}{person} & \rotatebox{60}{plant} & \rotatebox{60}{sktbrd} & \rotatebox{60}{train} & \rotatebox{60}{truck} & \rotatebox{60}{Avg} \\
		\hline
		\multirow{4}{*}{\rotatebox{30}{B$^2$UDA}}& Source Model \cite{dalib} & 76.4 & 26.4 & 60.2 & 80.2 & 73.5 & 2.3 & 89.4 & 8.2 & 79.7 & 25.6 & 69.4 & 3.8 & 49.6  \\
		& sMDA \cite{chidlovskii2016domain} & \textbf{77.8} & 39.3 & 66.1 & 73.7 & 74.3 & 4.2 & 87.9 & 16.7 & 79.8 & 36.9 & 71.4 & 9.8 & 53.1 \\
		& SoFA \cite{yeh2021sofa} &-- &--&--&--&--&--&-- &--&--&--&--&--& 60.4\\
		& IterLNL & 77.0 & \textbf{84.6} & \textbf{85.1} & \textbf{92.0} & \textbf{92.1} & \textbf{74.1} & \textbf{92.6} & \textbf{49.1} & \textbf{89.1} & \textbf{91.7} & \textbf{84.5} & \textbf{49.4} & \textbf{80.1} \\
		\hline
		\hline
		\multirow{2}{*}{\rotatebox{30}{WBUDA}}
		&3C-GAN \cite{Li_2020_CVPR} & 94.8 & 73.4 & 68.8 & 74.8 & 93.1 & 95.4 & 88.6 & 84.7 & 89.1 & 84.7 & 83.5 & 48.1 & 81.6 \\
		&SHOT \cite{liang2020we} & 94.3 & 88.5 & 80.1 & 57.3 & 93.1 & 94.9 & 80.7 & 80.3 & 91.5 & 89.1 & 86.3 & 58.2 & 82.9 \\
		\hline
		\hline
		\multirow{3}{*}{\rotatebox{30}{UDA}} & DANN \cite{dann} & 81.9 & 77.7 & 82.8 & 44.3 & 81.2 & 29.5 & 65.1 & 28.6 & 51.9 & 54.6 & 82.8 & 7.8 & 57.4 \\
		&MCD \cite{mcd} & 87.0 & 60.9 & 83.7 & 64.0 & 88.9 & 79.6 & 84.7 & 76.9 & 88.6 & 40.3 & 83.0 & 25.8 & 71.9 \\
		&RWOT \cite{xu2020reliable} & 95.1 & 80.3 & 83.7 & 90.0 & 92.4 & 68.0 & 92.5 & 82.2 & 87.9 & 78.4 & 90.4 & 68.2 & 84.0 \\
		\hline
	\end{tabular}
	\caption{Results on VisDA-2017 dataset, where all methods are based on a ResNet-101 model.}
	\label{Table:visda}
\end{table*}

\begin{figure}
	\centering
	\subfigure[Results with various $\gamma$] {
		\label{Fig:gamma}
		\includegraphics[width=0.225\textwidth]{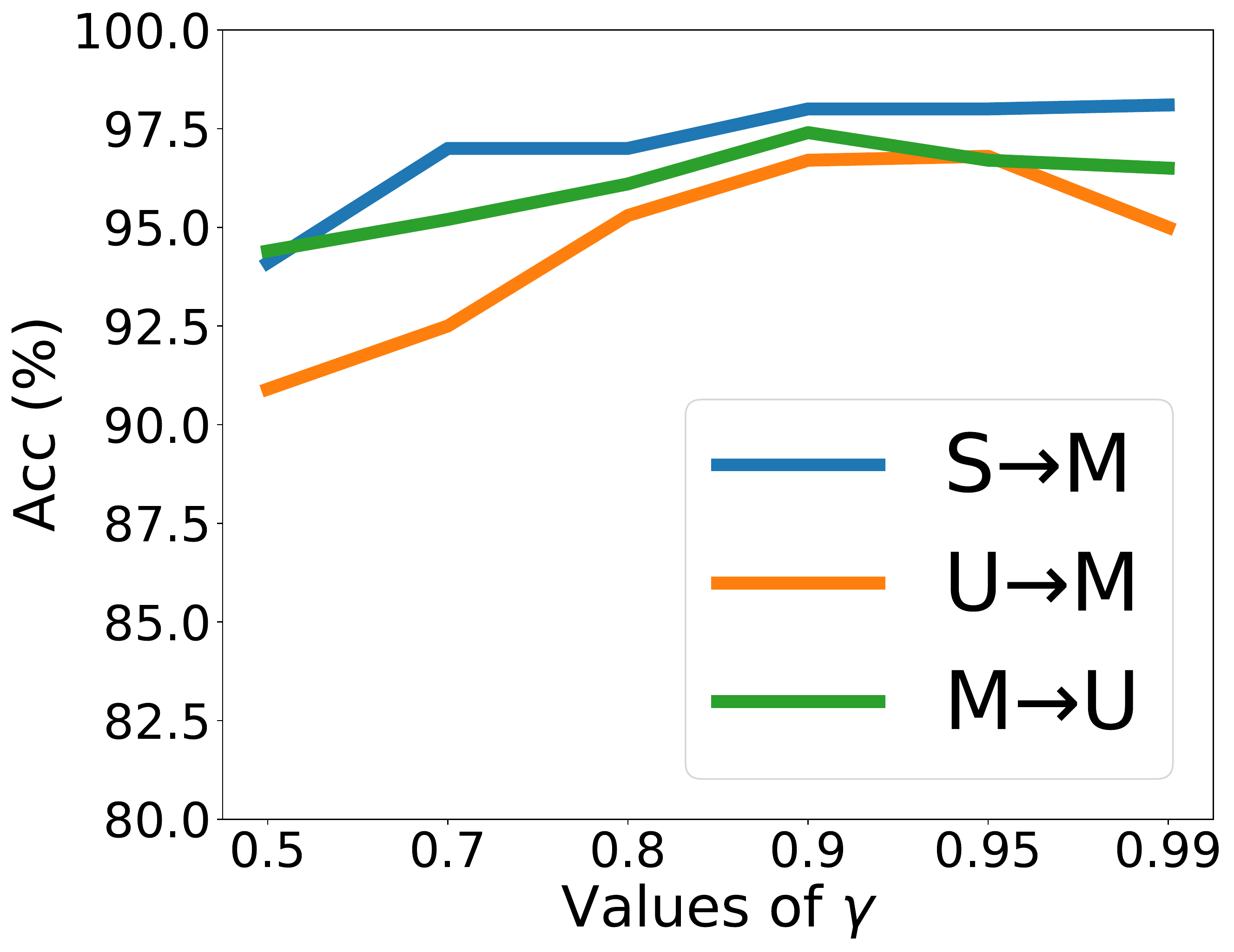}
	} \hfill
	\subfigure[Results with various $\kappa$] {
		\label{Fig:kappa}
		\includegraphics[width=0.225\textwidth]{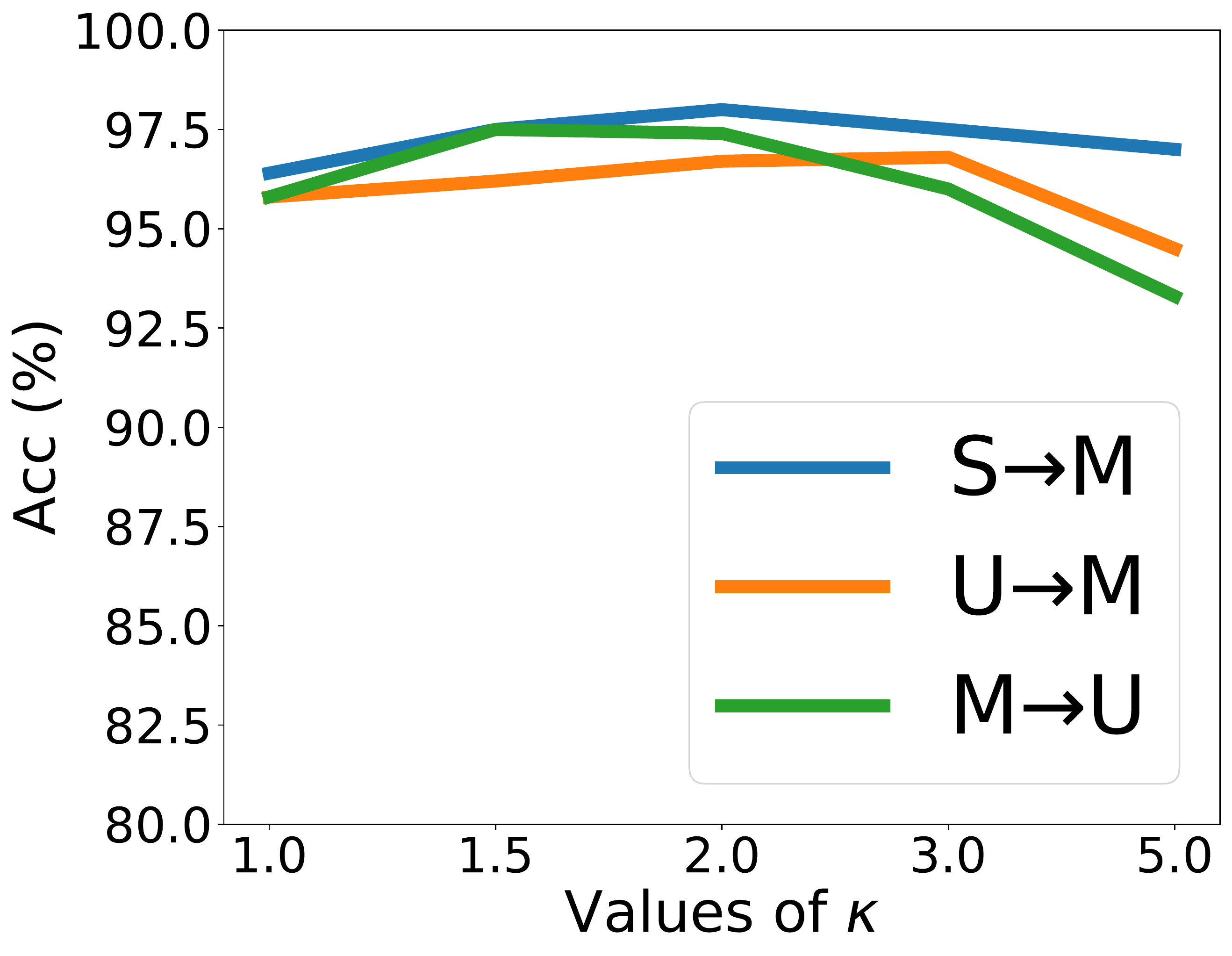}
	}
	\caption{IterLNL's  results with various values of $\gamma$ (\ref{Equ:proportion_high_confidence}) and $\kappa$ (\ref{Equ:rescale}).} \label{Fig:gamma_kappa}
	\vspace{-7pt}
\end{figure}

\subsection{Analysis}
\label{Sec:analysis}
\noindent \textbf{Analyses on $\gamma$ and $\kappa$.}
We investigate the hyper-parameters  $\gamma$ (\ref{Equ:proportion_high_confidence}) and $\kappa$ (\ref{Equ:rescale}) on the Digits datasets considering the experimental speed.
As illustrated in Figure \ref{Fig:gamma_kappa}, IterLNL performs robustly under a wide range of values. We adopt $\gamma=0.9$ and $\kappa=2.0$ in all experiments; such simple settings work well but not necessarily lead to the best results.
Similar results are also observed in the study of buffer length $h$ in Section \ref{Sec:LNL} and $n_k$ (\ref{Equ:sample_strategy}), as presented in the appendices. 

\noindent \textbf{Comparison with LNL Method.}
We conduct the LNL method Co-teaching \cite{han2018co} with our estimated noise rate $\epsilon$ (\ref{Equ:noise_level}) and noisy labels $\{ \widehat{y}_i^t\}_{i=1}^{n^t}$ introduced by the black-box source model $\widehat{F}^s$ on the S$\to$M task. The result of $91.0\%$ with Co-teaching is lower than $98.0\%$ with IterLNL, justifying the advantage of IterLNL over vanilla LNL method on B$^2$UDA.

\subsection{Results}

The results of IterLNL on datasets of Digits, Office31, and VisDA-2017 are illustrated in Table \ref{Table:digit}, Table \ref{Table:office31}, and Table \ref{Table:visda}, respectively. Most comparable results are directly reported from their original papers, except the Source Model \cite{dalib} and sMDA \cite{chidlovskii2016domain}, which are implemented by ourselves. Taking advantages of feature learning, our IterLNL improves over existing methods of B$^2$UDA \cite{chidlovskii2016domain,yeh2021sofa} significantly; for example, IterLNL improves over sMDA \cite{chidlovskii2016domain} and SoFA \cite{yeh2021sofa} by $27.0\%$ and $19.7\%$ on the VisDA-2017 dataset respectively; our IterLNL also improves over \cite{morerio2020generative}  by presenting good generalization on various benchmark datasets.
Although we only use the black-box source model for the transfer use in target domain, IterLNL achieves comparable results to methods of WBUDA, where the white-box source model is required, and traditional UDA, where labeled source data are required. 

\section{Conclusion and Broader Impact}

Considering that the source model itself may not be available due to commercial and/or safety considerations \cite{openai}, we investigate the B$^2$UDA task, and propose a baseline algorithm of IterLNL. We verify its efficacy on popular DA benchmarks with thorough experiments. 
The B$^2$UDA task is of broad practical value since it could further improve the utilities of APIs as cloud services, pushing the DA research closer to practical applications.


{\small
\bibliographystyle{ieee_fullname}
\bibliography{egbib}
}

\end{document}